\documentclass[runningheads]{llncs}
\usepackage[T1]{fontenc}
\usepackage{hyperref}
\usepackage{booktabs}
\usepackage{graphicx}
\usepackage[colorinlistoftodos,prependcaption,textsize=tiny]{todonotes}

\begin{document}
\title{A Deep Learning-Based Approach for Mangrove Monitoring}
\author{Lucas José Velôso de Souza\inst{1}
 \and
Ingrid Valverde Reis Zreik\inst{1}
\and
Adrien Salem-Sermanet\inst{1}
\and
 Nacéra Seghouani\inst{1,2}
 \and
Lionel Pourchier\inst{3}
 \institute{CentraleSupélec, Saclay Campus, Gif-sur-Yvette, 91190, France\\
 \email{\{ingrid.valverde-zreik, lucasjose.velosodesouza, adrien.salemsermanet\}@student-cs.fr}\\
 \and
LISN Paris-Sacaly University, Orsay, 91400, France\\
\email{nacera.seghouani@lisn.fr}\\
 \and 
EverSea -  \url{https://www.eversea.blue} \\ \email{ lionel.pourchier@eversea.blue }\\
 }
}
 \authorrunning{L. J. Velôso de Souza et \it{al}.}

\maketitle             
%
\begin{abstract}
Mangroves are dynamic coastal ecosystems that are crucial to environmental health, economic stability, and climate resilience. The monitoring and preservation of mangroves are of global importance, with remote sensing technologies playing a pivotal role in these efforts. The integration of cutting-edge artificial intelligence with satellite data opens new avenues for ecological monitoring, potentially revolutionizing conservation strategies at a time when the protection of natural resources is more crucial than ever. The objective of this work is to provide a comprehensive evaluation of recent deep-learning models on the task of mangrove segmentation. We first introduce and make available a novel open-source dataset, MagSet-2, incorporating mangrove annotations from the Global Mangrove Watch and satellite images from Sentinel-2, from mangrove positions all over the world. We then benchmark three architectural groups, namely convolutional, transformer, and mamba models, using the created dataset. The experimental outcomes further validate the deep learning community's interest in the Mamba model, which surpasses other architectures in all metrics.

\end{abstract}

\section{Introduction}

At the nexus of land and water, mangroves are vital ecosystems with significant environmental, economic, and societal benefits. They support marine biodiversity and fisheries, essential for coastal communities, while protecting shorelines from erosion and climate change impacts. Additionally, mangroves contribute to carbon sequestration and water purification, underscoring their role in environmental sustainability. The increasing degradation of these ecosystems poses significant risks to both the socio-economic stability of coastal areas and global environmental health, emphasizing the need for focused conservation efforts.

Globally, initiatives and partnerships \footnote{
\href{https://www.worldwildlife.org/initiatives/mangroves-for-community-and-climate}{World Wildlife Fund (WWF), ``Mangroves for Community and Climate''.} } are advancing efforts to safeguard mangrove ecosystems, supported by legal and policy frameworks tailored to international conservation goals. These endeavors combine science, community action, and policy innovation to ensure the future of mangrove habitats. 
Building upon this foundation, the collection and analysis of accurate information on areas affected by degradation and vegetation changes become indispensable. This data is crucial for effective decision-making in environmental management and conservation investment, serving as a cornerstone for the strategies and actions developed to protect these vital ecosystems. Remote sensing technologies allow both a large scale monitoring and precise quantification of changes in vegetation cover, facilitating informed decisions on resource allocation for conservation and restoration efforts. 
Technological advancements in remote sensing and artificial intelligence offer promising avenues for monitoring and protecting these ecosystems. For instance, previous work have shown that the application of Convolutional Neural Networks (CNNs) in mangrove analysis supports detailed ecological assessments and strategic decision-making in mangrove management~\cite{Iovan2020}. However, with the recent advancements and enhancements in the field of Deep Learning, such as the development of Transformer models \cite{vaswani2023attention} and recently Mamba techniques \cite{gu2023mamba}, the use of new and better performing methodologies becomes crucial for achieving more accurate outcomes and to better inform decision-makers.

Building upon these innovative approaches for improved mangrove segmentation and monitoring, the integration of geolocation data with updated satellite imagery is key to unlocking new potentials in ecosystem conservation. Despite the vast availability of data today, effectively accessing mangroves positions for the observation of this ecosystem remains a challenge. Although many organizations have detailed mangrove location data, it is often not integrated with satellite imagery, a crucial step for effective ecosystem management. This paper introduces a new dataset, \textit{MagSet-2}, which combines mangrove geolocation data from \textit{Global Mangrove Watch} ~\cite {rs14153657} with \textit{Sentinel-2} satellite images ~\cite{sentinelhub2024}. This integration catalyzes machine learning models that are vital for accurately monitoring and predicting mangrove locations. Furthermore, we concentrate our efforts in comparing various deep learning architectures to interpret satellite imagery, namely CNN-based architectures (U-Net\cite{ronneberger2015u}, MANet\cite{Zhang2018MANet}, PAN\cite{Zhao2018PAN}), Transformer-based architectures (BEiT\cite{bao2021beit}, Segformer \cite{xie2021segformer}) and a Mamba-based architecture (Swin-UMamba\cite{liu2024swinumamba}). These architectures are chosen due to their recognized effectiveness in semantic segmentation tasks. By developing a comprehensive dataset and employing a novel Mamba-type model, this research aims to enhance the precision of mangrove ecosystem analysis, supporting more informed strategies.

In the following, we delve into a  review of related work in the field of mangrove detection and segmentation in Section \ref{sec:rw}. Section \ref{sec:overview} offers insights into the dataset creation process, detailing how mangrove annotations were collected and how Sentinel-2 imagery was acquired, which are essential components for our research \footnote{GitHub with Code and Dataset: \url{https://github.com/SVJLucas/MangroveAI}}. Section \ref{sec:metho} presents our approach to mangrove segmentation, including the implementation of deep learning models and the utilization of advanced techniques such a the Mamba-type architecture. Continuing, Section \ref{sec:exp} thoroughly evaluates the six leading deep learning architectures for mangrove segmentation, highlighting their performance and suitability for the task. Finally, Section \ref{sec:conclu} summarizes the evaluation results.

\section{Related Work}
\label{sec:rw}
Historically, mangrove monitoring has evolved significantly, driven by advancements in remote sensing technology, computational methods, and machine learning techniques. Initially, mangrove monitoring relied on field surveys, which were labor-intensive and limited by the challenging and inaccessible nature of mangrove habitats \cite{Anh2022}. With the advent of remote sensing, researchers gained the ability to observe mangroves over extensive areas with greater efficiency and reduced cost. Early remote sensing methods employed basic vegetation indices like NDVI, EVI, and VARI to assess the health and distribution of mangroves \cite{Houborg2007}.

As machine learning gained prominence, both unsupervised and supervised algorithms began to be integrated into remote sensing workflows. Techniques like K-nearest neighbors (KNN) \cite{Woltz2022}, support vector machines (SVMs) \cite{Vidhya2014}, and random forests (RF)  \cite{Behera2021} were applied to improve the accuracy of mangrove classification based on spectral data. However, these methods often struggled with spectral ambiguities, where different objects share similar spectral signatures, leading to misclassification \cite{He2020}.

To overcome the limitations of single-method approaches, researchers started combining traditional index methods with machine learning algorithms. This hybrid approach leveraged the strengths of both techniques, using indices to guide machine learning models, thereby enhancing the accuracy of mangrove extraction and analysis \cite{ValderramaLanderos2021}.

The breakthrough of deep learning in image recognition, marked by the success of AlexNet in 2012 \cite{Krizhevsky2017}, paved the way for sophisticated architectures like CNNs, namely FCN , SegNet~\cite{badrinarayanan2016segnet} , and U-Net~\cite{ronneberger2015u} for semantic segmentation of remote sensing images \cite{Guo2021,Long2017}. These deep learning models have been particularly transformative in mangrove monitoring, offering significant improvements in segmentation accuracy through their ability to learn complex patterns and features from large datasets. However, a notable challenge persists as most of these advancements are constrained by geographical and ecological specificity, primarily due to the absence of a comprehensive global dataset that maps diverse mangrove species across various regions.

The complexity of mangrove ecosystems and the diverse challenges in their monitoring, thereby underscores the pressing need for a globally representative mangrove dataset to further enhance the effectiveness and scalability of deep learning applications in mangrove monitoring. By comparing the CNNs, Transformers and Mamba architectures, we aim to identify the most effective deep learning model for mangrove segmentation. This comparison is critical not only for enhancing the accuracy of current monitoring efforts but also for pushing the boundaries of what can be achieved in ecological monitoring and conservation with state-of-the-art technology.

\section{Mangrove Data}
\label{sec:overview}
 We aimed to build and release MagSet-2
 of mangrove regions worldwide  to perform benchmarking analyses and serving as the foundation for training the proposed models. This dataset consists of annotations of different species of mangroves from the \textit{Global Mangrove Watch} dataset~\cite {rs14153657} and corresponding \textit{Sentinel-2} optical satellite imagery~\cite{sentinelhub2024}. The building of this dataset answers to the non-availability of a global heterogeneous mangrove annotated dataset.

\begin{figure}[t]
    \centering
    \includegraphics[width=0.9\linewidth]{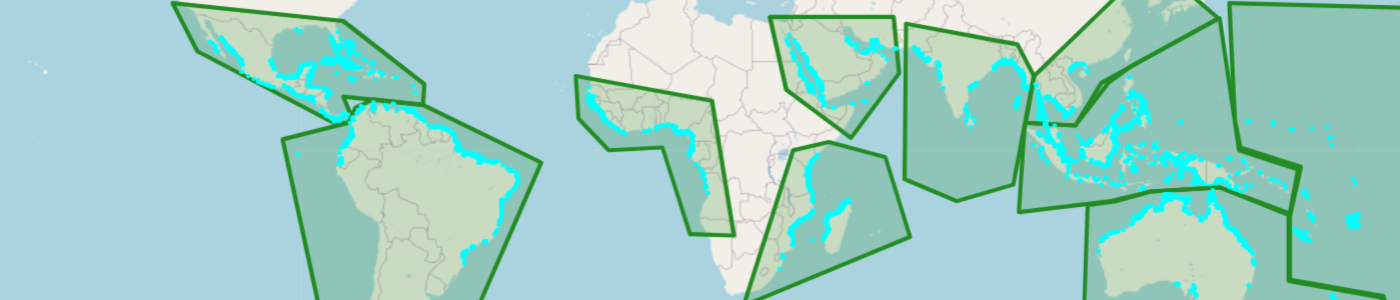}
    \caption{Mangrove Position (neon blue) and the different Mangrove Zones (green) Dataset based on the Global Mangrove Watch (GMW) v3.2020}
    \label{fig:gmw_zones_and_polygons}
\end{figure}

\begin{figure*}[t]
    \centering
\includegraphics[width=0.8\linewidth]{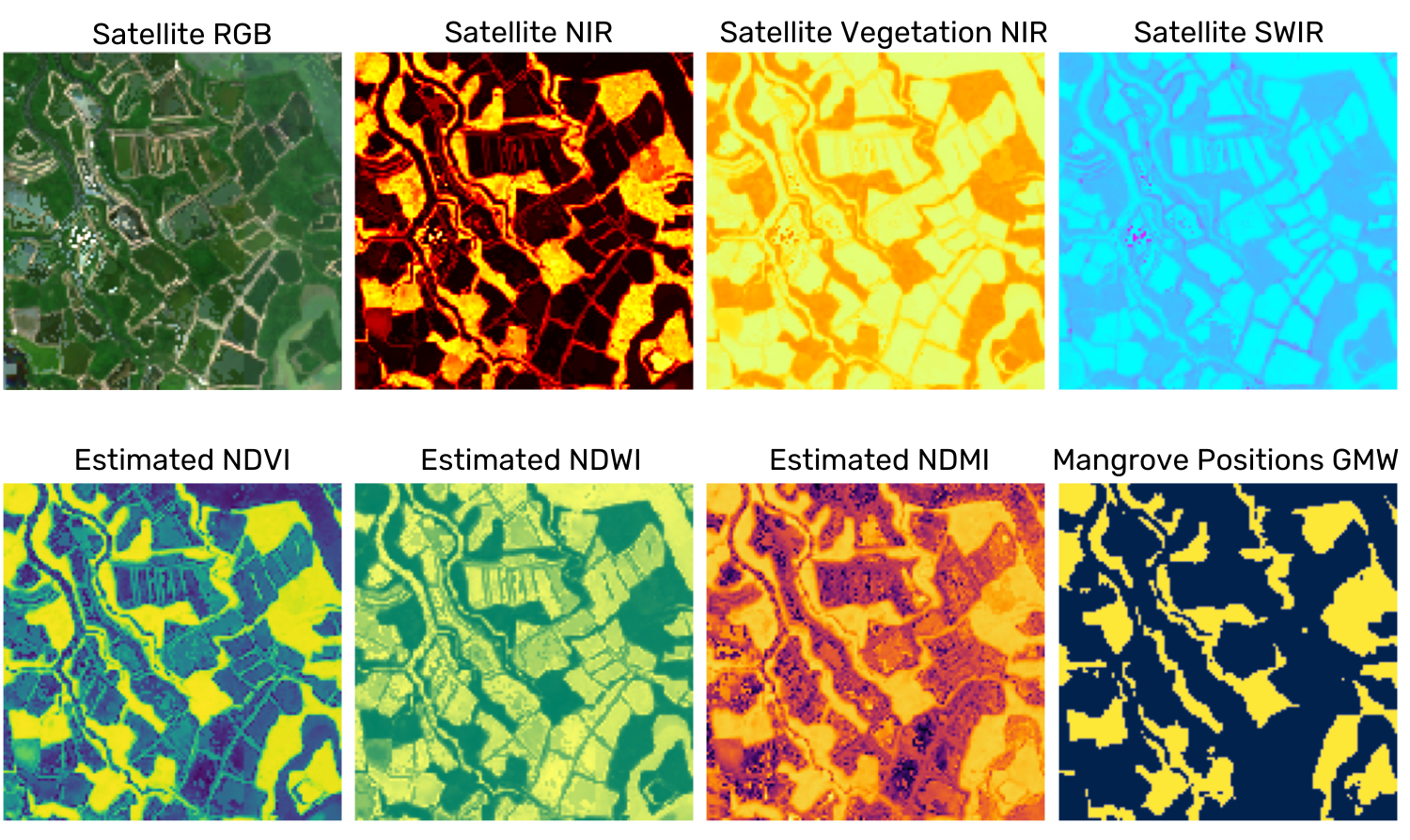} 
    \caption{Sentinel-2 Spectral Display and Vegetation Analysis: Starting from the top left with the RGB bands, followed by the NIR band, Vegetation NIR, and SWIR band in sequence. On the bottom row, from left to right, we have the estimated NDVI, NDWI, NDMI indices, and the targeted Mangrove locations for predictive modeling.} 
    \label{fig:plotcomparison9} 
\end{figure*}

The \textit{Global Mangrove Watch} dataset of annotations~\cite {rs14153657}, as well as a world map of mangrove distribution~\cite{mangrove_distribution}, were used to define $10$ non-overlapping geographic zones: Central America, South America, West Africa, East Africa, Middle East, India, East Asia, Indonesia, Australia and Pacific Islands. These zones, shown in Figure \ref{fig:gmw_zones_and_polygons}, depict geographic regions with distinct mangrove characteristics to facilitate stratified sampling and improve model performance across diverse environments. 
\textit{Sentinel-2} satellites provide multispectral imagery~\cite{sentinelhub2024} for various Earth observation applications. Since mangrove reflectance was observed to rise rapidly towards the red edge~\cite{rs14194868}, mangrove ecosystems can be observed using indices computed from spectral bands in the visible and infrared regions of optical remote sensing. That is why the Sentinel-2 bands used in training include $B$ (Blue), $G$ (Green), $R$ (Red), $NIR$ (Near-Infrared), $NIR_{vegetation}$ (Vegetation Red Edge) and $SWIR$ (Short-Wave Infrared), as can be observed in Figure \ref{fig:plotcomparison9}. 

From these spectral bands, we were able to derive the following relevant spectral indices used in mangrove segmentation~\cite{rs14194868}:  NDVI (Normalized Difference Vegetation Index), NDWI (Normalized Difference Water Index), and NDMI (Normalized Difference Moisture Index). These indices were computed using the formulas in Equations \ref{eq:ndvi}, \ref{eq:ndwi} and \ref{eq:ndmi} and used as additional input channels.

\begin{equation}
\scriptsize
    NDVI = \frac{{NIR - R}}{{NIR + R}}
    \label{eq:ndvi}
\end{equation}

\begin{equation}
\scriptsize
    NDWI = \frac{{G - NIR}}{{G + NIR}}
    \label{eq:ndwi}
\end{equation}

\begin{equation}
\scriptsize
    NDMI = \frac{{NIR - SWIR}}{{NIR + SWIR}}
    \label{eq:ndmi}
\end{equation}

\noindent In the context of Sentinel-2 data~\cite{sentinelhub2024}, preprocessing levels indicate the amount of calibration and processing applied to the raw sensor data. The level used in our work is the Level-2A, which includes radiometric, geometric, and atmospheric corrections.

Particular emphasis was placed on addressing the prevalent issue of cloud coverage. Globally, cloud cover averages approximately 70\%, with even higher percentages over oceans and near the equator regions that host the majority of the world's mangrove ecosystems \cite{essd-14-5573-2022}. Given that optical sensors are inherently passive, clouds can obstruct the terrestrial view in satellite imagery, either partially or completely, thereby hindering precise segmentation. To mitigate this, we employed a technique involving the creation of "cloudless mosaics" at the time of obtaining \textit{Sentinel-2} images. This technique entails composing an image from the clearest pixels of multiple images captured at the same location over a predefined period \cite{PITKANEN2024103659}. We selected the time frame from March to June 2020. Ultimately, we set the standard image resolution $(W\times H)$ at $128 \times 128$, subdividing larger acquisitions into several sub-images to maintain this resolution. The final dataset, MagSet-2, comprises 10,483 pairs of spectral bands and indices, along with mangrove location annotations.

\section{Methodology}
\label{sec:metho}

Due to hardware limitations, our benchmarking process was conducted in two phases. In the first phase, we evaluated different models from distinct categories—including convolutional, transformers, and mamba—using a reduced sample of the MagSet-2 dataset. In the second phase, the top-performing models from each category were selected for a comprehensive evaluation, where they were trained and tested on the full dataset.

For the initial evaluation stage, we began by selecting a 5\% sample of the MagSet-2 dataset. This sample was equally divided into training and testing subsets, with each comprising 50\% of the data. This training data is meticulously prepared for robust mangrove segmentation through a series of strategic preparation steps: \textbf{(i)} Initiation involves geographic shuffling across zones to eliminate location-based biases and promote diverse sample representation, essential for the model to generalize across varying environments. \textbf{(ii)} Preprocessing includes data augmentation (vertical/horizontal flips, 90-degree rotations, random resizing crops) to introduce variability, and \textbf{(iii)} normalization of image pixels to a standardized range (0-1) to address disparities in brightness and contrast. \textbf{(iv)} The introduction of Gaussian noise simulates real-world disturbances, enhancing the model's generalization capabilities. These steps culminate in a well-structured training dataset, ready for effective training and evaluation within the deep learning paradigm.

Following the preprocessing of the sampled dataset, we assess the performance of six state-of-the-art deep learning architectures: U-Net \cite{ronneberger2015u}, MANet \cite{Zhang2018MANet}, PAN \cite{Zhao2018PAN}, BEiT \cite{bao2021beit}, Segformer \cite{xie2021segformer}, and Swin-UMamba \cite{liu2024swinumamba}, within the context of mangrove segmentation. These architectures were chosen based on their prominence and demonstrated effectiveness in semantic segmentation tasks.

After identifying the most promising model from each category—convolutional, transformers, and mamba—we utilized the complete MagSet-2 dataset for further evaluation. The full dataset was similarly divided into a training set (50\%) and a testing set (50\%), with the same preprocessing steps applied. Finally, we retrained the three best-performing models on this data.

This comparative analysis serves as a pivotal step towards identifying the most suitable architecture for mangrove segmentation applications, thereby advancing the state-of-the-art techniques in environmental monitoring and conservation efforts.

\section{Experiments and Evaluation}

\subsection{Sampled MagSet-2 Dataset Evaluation}

\label{sec:exp}
We utilized convolutional architectures such as U-Net \cite{ronneberger2015u}, MANet \cite{Zhang2018MANet}, and PAN \cite{Zhao2018PAN}, all integrated with a ResNet50 encoder \cite{he2016deep} pretrained on ImageNet \cite{russakovsky2015imagenet}. These differ from BEiT \cite{bao2021beit}, which employs transformer blocks for feature extraction, and SegFormer \cite{xie2021segformer}, which uses a transformer encoder paired with a lightweight decoder for semantic segmentation. Swin-UMamba \cite{liu2024swinumamba} relies on the Swin-Transformer \cite{liu2021swin} architecture as its encoder.

All models employed a sigmoid activation function. Hyperparameters included a uniform batch size of 32 across models, with convolutional models set at a learning rate of $1 \times 10^{-4}$ and both transformer and the mamba models at $5 \times 10^{-4}$. Training extended over 100 epochs, using the AdamW optimizer \cite{loshchilov2019decoupled} with a learning rate scheduling mechanism that halves the rate after every 7 epochs without improvement. The loss function used was Binary Cross-Entropy.

Our experimental design aimed to ensure a fair comparison by maintaining a consistent number of parameters across models to control for computational complexity; each model had approximately $33$M parameters. The analysis was conducted using a single NVIDIA A100 40 GB GPU. The output from each model was a segmentation map in a single channel, indicating the probability of each pixel belonging to the mangrove class.

\begin{figure*}[h!]
    \centering   \includegraphics[width=\linewidth]{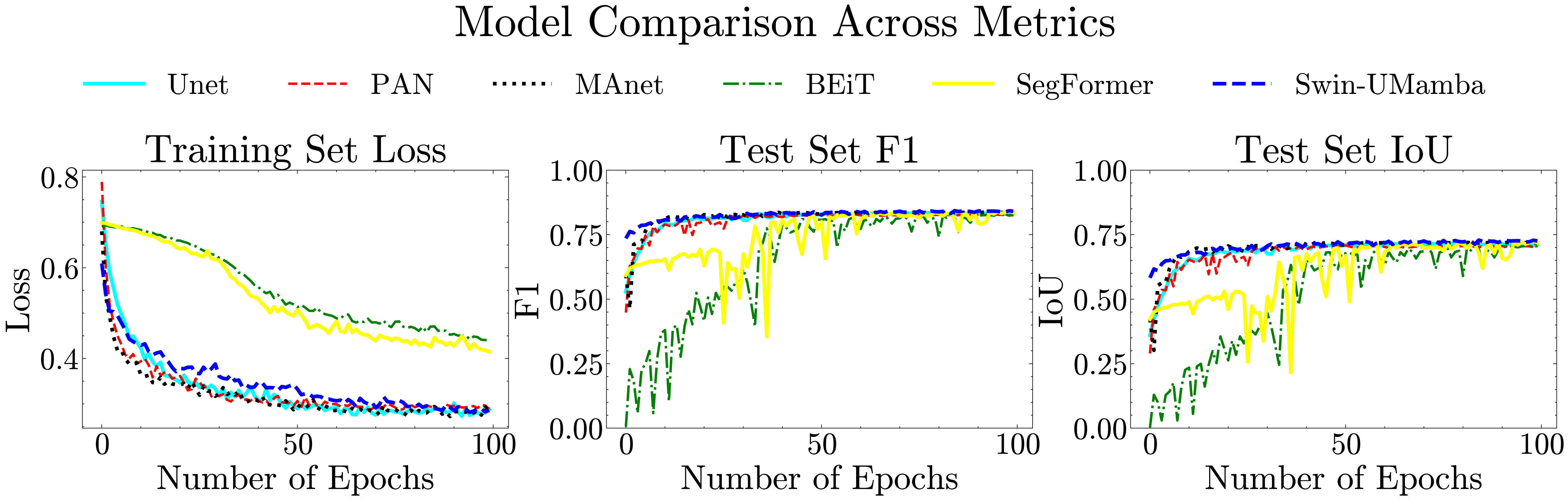}   
    \caption{Comparative performance on sampled test of MagSet-2,  using Training Set Loss (left), Test Set F1 Score (center), and Test Set Intersection over Union (IoU) (right). Each line represents a model: U-Net (neon blue), PAN (red), MANet (black), BEiT (green), SegFormer (yellow), and Swin-UMamba (dark blue) trained over 100 epochs. Lower loss values, higher F1 and IoU values indicate better performance. Swin-UMamba consistently shows superior performance over all metrics.}
    \label{fig:comp_perf}
\end{figure*}

We assess the performance of each model using established image segmentation evaluation metrics, complemented by a qualitative analysis of the results. We focus on the Intersection over Union (IoU), also known as the Jaccard Index, which measures the extent of overlap between the predicted masks and the actual ground truth. This metric is crucial for evaluating the precision of the segmentation, as it quantifies the relative error by considering both the area of overlap and the total area covered by both the predicted and actual masks. Other straightforward metrics used include Accuracy, which gauges the proportion of correctly classified samples, the F1-score, balancing precision and recall, and Loss, Binary Cross-Entropy, which helps in optimizing the training process.
The performance of the selected deep learning models for mangrove segmentation on \textit{Sentinel-2} satellite imagery is presented in Table \ref{tab:tb_results}. The six models, categorized into three architectural groups, were evaluated:
\begin{itemize}
    \item \textbf{Convolutional models}: U-Net, PAN, and MAnet achieved moderate performance, with U-Net having the lowest number of parameters (32.54 million) but also the lowest IoU score (61.76\%). MAnet achieved the lowest Test Loss (0.34) among the convolutional ones, while PAN stood in the middle ground between the two.
    \item \textbf{Transformer models}: BEiT and SegFormer, both utilizing transformer architectures, showed improved performance compared to convolutional models. SegFormer achieved the highest IoU score (72.32\%) within this group, suggesting its effectiveness in capturing complex spatial relationships in mangrove imagery.
    \item \textbf{Mamba model}: Swin-UMamba surpassed all other models in all metrics. It achieved the highest IoU (72.87\%), Accuracy (86.64\%), F1-score (84.27\%), and the lowest Loss (0.31). This suggests Swin-UMamba's capability to learn efficient and accurate representations of mangroves in \textit{Sentinel-2} data.
\end{itemize}
\begin{table*}[t]
  \centering
  \begin{tabular}{lcccccc}
    \toprule
    & \multicolumn{1}{c}{\# Parameters (M)} & \multicolumn{4}{c}{Performance Metrics on MagSet-2} \\
     \cmidrule(lr){3-6}
    Method &  & IoU$\uparrow$ (\%) & Acc$\uparrow$ (\%) & F1-score$\uparrow$ (\%) & Loss$\downarrow$ \\
    \midrule
    U-Net & 32.54 & 61.76 & 78.59 & 76.32 & 0.47 \\
    PAN & 34.79 & 64.44 & 81.16 & 78.32 & 0.41 \\
    MAnet & 33.38 & 71.75 & 85.80 & 83.51 &  0.34 \\
    \midrule
    BEiT & 33.59 & 70.78 & 85.66 & 82.87 &  0.48 \\
    SegFormer & 34.63 & 72.32 & 86.13 & 83.91 &  0.42 \\
    \midrule
    Swin-UMamba & 32.35 & \textbf{72.87} & \textbf{86.64} & \textbf{84.27} & \textbf{ 0.31} \\
    \bottomrule
  \end{tabular}
  \caption{Performance Comparison on Sampled MagSet-2 Dataset.}
  \label{tab:tb_results}
\end{table*}
The Figure \ref{fig:comp_perf} presents the comparative analysis across the loss, F1-score and IoU metrics tracked over a training period of 100 epochs. In this context, to evaluating model performance on unseen data, the F1-score is crucial for understanding precision and recall balance (Figure \ref{fig:comp_perf}, center). Swin-UMamba's performance plateaued at the highest F1-score, reflecting consistent and superior segmentation accuracy. The remaining models displayed improvement over the training epochs, with SegFormer trailing closely behind Swin-UMamba, signifying the effectiveness of transformer-based architectures in handling the complexity of mangrove segmentation.

\begin{figure}[h!]
    \centering    \includegraphics[width=0.9\linewidth]{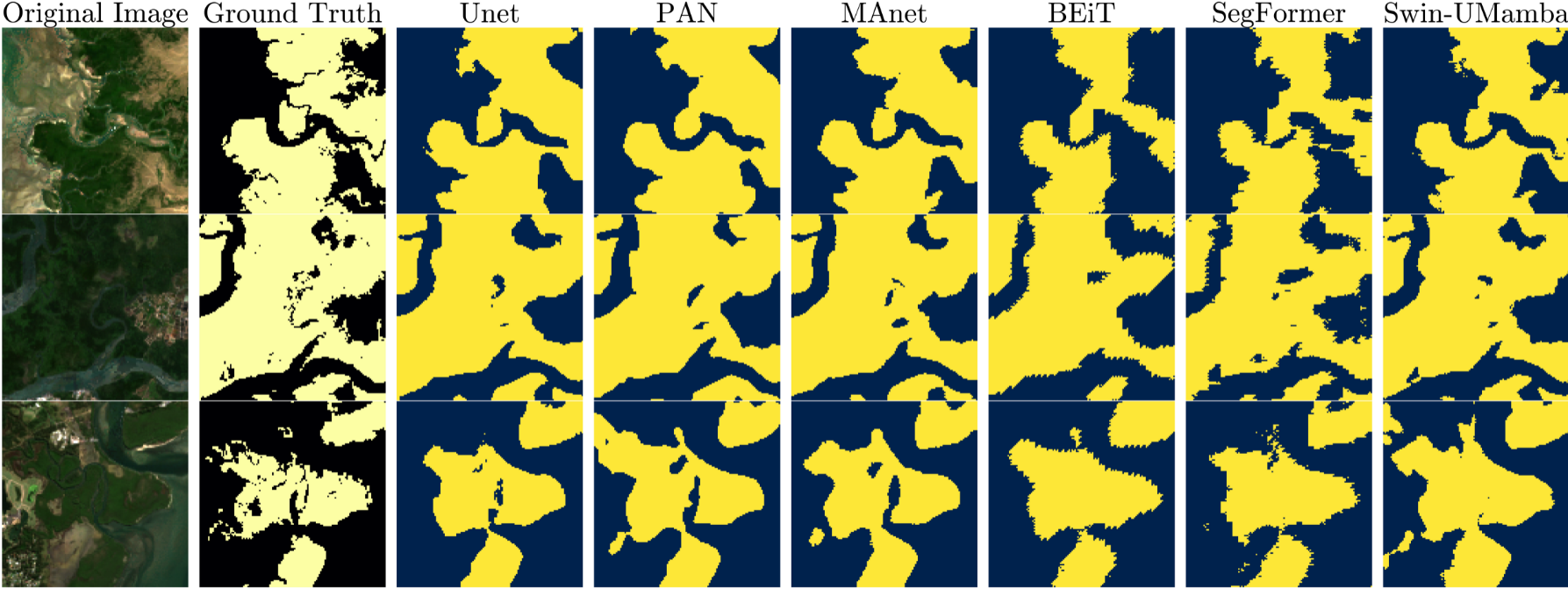}    
    \caption{Comparative visual segmentation results of mangrove areas. The first column shows the original satellite images, the second column depicts the ground truth segmentation, and the subsequent columns display the segmentation results from U-Net, PAN, MANet, BEiT, SegFormer, and Swin-UMamba  models, resp.}    \label{fig:models_comp}
\end{figure}

The Figure \ref{fig:models_comp} illustrates a side-by-side comparison of segmentation results for mangrove areas from satellite imagery. The visual comparison suggests that the convolutional group generates predictions with limited granularity and detail, compared to the ground-truth, whereas transformer models produce excessively noisy Mangrove prediction masks with overly high levels of detail, leading to increased errors. However, Swin-UMamba offers a superior balance between detail and prediction accuracy, delivering predictions that are more detailed than those from convolutional models, without indicating overfitting. These findings suggest that while convolutional models like U-Net offer reasonable performance with fewer parameters, transformer models like SegFormer and the newly introduced Swin-UMamba demonstrate superior capabilities in capturing the intricate characteristics of mangroves in satellite imagery. Notably, Swin-UMamba's superior performance across all metrics indicates its potential as a promising approach for accurate and efficient mangrove segmentation tasks.

\vspace*{-0.55cm}
\vspace*{-1cm}

\subsection{Complete MagSet-2 Dataset Evaluation}

In the final evaluation, using the full version of the MagSet-2 dataset, we performed a comprehensive benchmark of the top-performing models across three categories: convolutional-based (MAnet), transformer-based (SegFormer), and Mamba-based (Swin-UMamba). The models were evaluated under the same training configurations as in the previous experiments, but with the complete dataset. The results, summarized in Table \ref{tab:tb_results_full}, indicate that Swin-UMamba was the top performer across all metrics, achieving an Intersection over Union (IoU) of 76.90\% and an F1-score of 86.91\%, while maintaining the lowest loss value of 0.28. These results further highlight Swin-UMamba's robustness and adaptability in handling MagSet-2 the dataset with high precision.

\begin{table*}[t]
  \centering
  \begin{tabular}{lcccccc}
    \toprule
    & \multicolumn{1}{c}{\# Parameters (M)} & \multicolumn{4}{c}{Performance Metrics on MagSet-2} \\
    \cmidrule(lr){3-6}
    Method & & IoU$\uparrow$ (\%) & Acc$\uparrow$ (\%) & F1-score$\uparrow$ (\%) & Loss$\downarrow$ \\
    \midrule
    MAnet & 33.38 & 76.60 & 87.71 & 86.71 & 0.29 \\
    SegFormer & 34.63 & 75.37 & 86.61 & 85.92 & 0.35 \\
    Swin-UMamba & 32.35 & \textbf{76.90} & \textbf{87.75} & \textbf{86.91} & \textbf{0.28} \\
    \bottomrule
  \end{tabular}
  \caption{Performance Comparison on Full MagSet-2 Dataset.}
  \label{tab:tb_results_full}
\end{table*}

The MAnet model also exhibited strong performance, closely matching Swin-UMamba with an IoU of 76.60\% and an F1-score of 86.71\%. MAnet's ability to efficiently manage a large dataset underscores the continued relevance of convolutional architectures in segmentation tasks. By contrast, SegFormer, which initially demonstrated superior performance during the subsample evaluations, experienced a modest decline when applied to the full dataset, achieving an IoU of 75.37\% and an F1-score of 85.92\%. This change in relative performance between SegFormer and MAnet suggests that SegFormer, despite the typical advantage of transformer-based models on larger datasets, may not have fully realized its potential in this scenario.

\begin{figure*}[h!]
    \centering   \includegraphics[width=\linewidth]{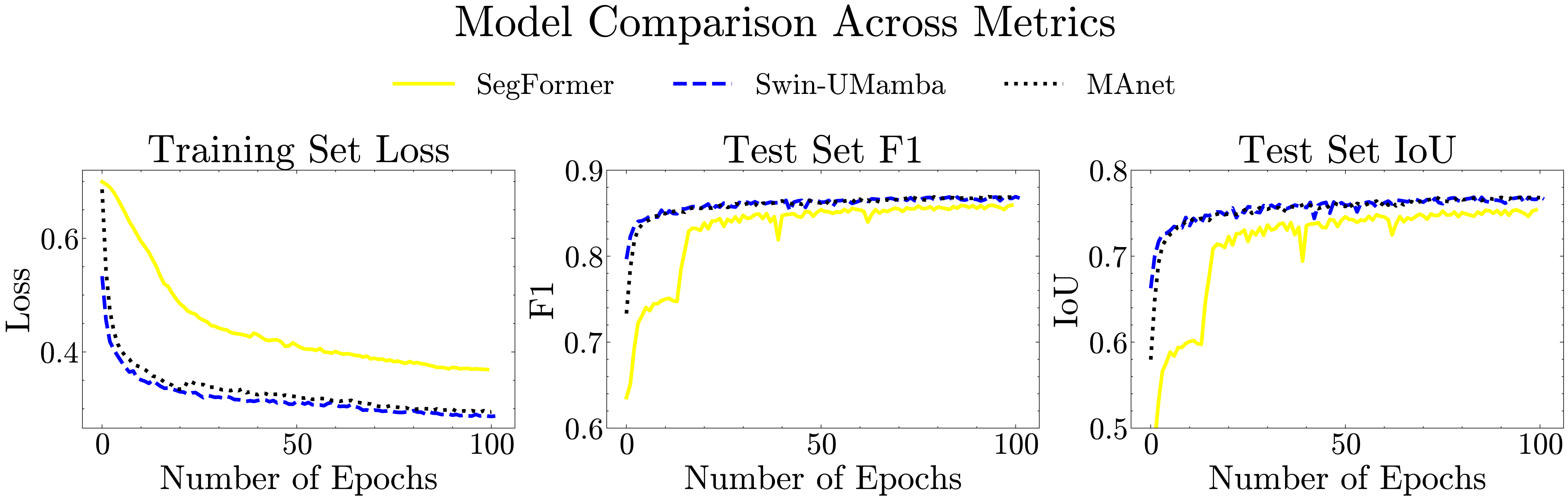}   
    \caption{Comparative performance on complete MagSet-2,  using Training Set Loss (left), Test Set F1 Score (center), and Test Set Intersection over Union (IoU) (right). Each line represents a model:  MANet (black), SegFormer (yellow), and Swin-UMamba (dark blue) trained over 100 epochs. Lower loss values, higher F1 and IoU values indicate better performance. Again, Swin-UMamba consistently shows superior performance over all metrics.}
    \label{fig:final_comp_image_full}
\end{figure*}

As shown in Figure \ref{fig:final_comp_image_full}, both the central and right-hand charts, which represent the F1-score and IoU respectively, indicate that the SegFormer curve continues to rise. This suggests that the model did not fully converge within the 100 epochs established in the experiment. Transformer-based models, such as SegFormer, generally require more time to converge compared to convolutional models \cite{xiao2021earlyconvolutionshelptransformers}. In this context, it is plausible that SegFormer required more than the 100 epochs used in this experiment to reach its optimal performance. This reinforces the notion that, although transformer models are often more suitable for larger datasets, SegFormer might have needed additional epochs or further hyperparameter tuning to improve generalization. Thus, while transformer-based models offer significant potential, they may require careful adjustment to prevent underperformance, particularly when working with larger and more complex datasets.

\vspace*{-0.55cm}

\section{Conclusion}
\label{sec:conclu}
 This study embarked on a comprehensive evaluation of various deep learning models for the segmentation of mangroves from satellite imagery, focusing on six architectures: U-Net, MANet, PAN, BEiT, SegFormer, and Swin-UMamba. This comparative analysis revealed that the Swin-UMamba model notably outperforms established convolutional and transformer architectures. Demonstrating superior performances and generalization in semantic segmentation, Swin-UMamba's advancements offer significant implications for ecological monitoring and the conservation of mangrove ecosystems. These results affirm the potential of leveraging state-of-the-art deep learning techniques in environmental remote sensing, providing a robust foundation for  enhancing our understanding the dynamics of these critical habitats on a global scale.  

\section{Acknowledgments}

We would like to express our sincere gratitude to \textit{Céline Hudelot} from the \textit{Laboratoire Mathématiques et Informatique pour la Complexité et les Systèmes (MICS)}, from \textit{École CentraleSupélec - Paris-Saclay University},  for her invaluable support and guidance throughout this work. Besides, this study was financed in part by the \textit{Coordenação de Aperfeiçoamento de Pessoal de Nível Superior – Brasil (CAPES)} – Finance Code 001.

\bibliographystyle{splncs04}
\bibliography{ref}

\end{document}